\documentclass[journal]{IEEEtran}

\usepackage[colorlinks,urlcolor=blue,linkcolor=blue,citecolor=blue]{hyperref}

\usepackage{color,array}

\usepackage{graphicx}
\usepackage{subfig}
\usepackage{amsmath}
\usepackage{amsfonts}
\usepackage{tablefootnote}
\usepackage{multirow}
\usepackage{multicol}
\usepackage{array}
\newcommand{\subfigANDtitle}[2][.2\linewidth]{%
  \begin{tabular}{@{}>{\centering\arraybackslash}p{#1}@{}} #2 \end{tabular}}
  
\begin{document}

\title{Separate Scene Text Detector for Unseen Scripts is Not All You Need}

\author{Prateek Keserwani$^1$, Taveena Lotey$^1$, Rohit Keshari$^2$, and Partha Pratim Roy$^1$\\
$^1$ Indian Institute of Technology Roorkee, India \\$^2$ Indraprastha Institute of Technology Delhi, India\\
\{pkeserwani, taveena\}@cs.iitr.ac.in, rohitk@iiitd.ac.in, partha@iitr.ac.in
} 
\markboth{Arxiv Preprint }
{}

\maketitle

\begin{abstract}
Text detection in the wild is a well-known problem that becomes more challenging while handling multiple scripts. In the last decade, some scripts have gained the attention of the research community and achieved good detection performance. However, many scripts are low-resourced for training deep learning-based scene text detectors. It raises a critical question: \emph{Is there a need for separate training for new scripts?} It is an unexplored query in the field of scene text detection. This paper acknowledges this problem and proposes a solution to detect scripts not present during training. In this work, the analysis has been performed to understand cross-script text detection, i.e., trained on one and tested on another. We found that the identical nature of text annotation (word-level/line-level) is crucial for better cross-script text detection. The different nature of text annotation between scripts degrades cross-script text detection performance. Additionally, for unseen script detection, the proposed solution utilizes vector embedding to map the stroke information of text corresponding to the script category. The proposed method is validated with a well-known multi-lingual scene text dataset under a zero-shot setting. The results show the potential of the proposed method for unseen script detection in natural images.
\end{abstract}


\section{Introduction}

\IEEEPARstart{T}{he} writing system is one of the elemental inventions of human ancestors, which is a necessary component of the current civilized world. The writing system evolved in different human civilizations and arose as multiple scripts. With trade and migration of humans from one culture to another, the scripts got influenced by each other. It led to the environment of multiple scripts in the same geographical location. A country such as India has eleven official scripts \cite{obaidullah2019automatic}. These days, the multi-script text is not confined to the written content of literature and science. It is also present on the signboards, posters, t-shirts, shop front, packed products, etc. These written content is known as scene text \cite{ye2014text,liu2019scene,zhu2021textmountain}. These scene texts may contain text instances of multiple scripts. Localizing \cite{cai2020spn,keserwani2021quadbox} and recognizing multiple scripts \cite{mei2016scene,gomez2017improving,bhunia2019script,shi2015automatic,gomez2016fine} in natural images is a substantial problem.

In the last decade, many state-of-the-art methods have been proposed for multi-script scene text detection \cite{raghunandan2018multi,saha2020multi}. All the existing methods have assumed that the targeted script has an abundance of well-labeled datasets for the training purpose of deep learning algorithms. Many countries have several scripts, and among them, some are low-resourced. Therefore, adding new scripts to the existing scene text detector needs additional training with new scripts. Also, for the case of low-resourced scripts, the annotated training data is an additional challenge. Hence, in place of separate training for each script, it is beneficial to devise an approach that detects the text for unseen scripts with the help of a one-time trained model. The proposed work considers the point mentioned above as a motivation. This approach helps to eliminate the unnecessary burden of separate training for unseen scripts (i.e., zero-shot setting).

In object detection, the above-mentioned scenario, i.e., detection of the unseen object category while testing, has been analyzed, and various works are available in the literature \cite{bansal2018zero,rahman2018zero,li2019zero}. This setting of testing is known as zero-shot setting. However, text detection under the zero-shot setting still needs to be explored. All these existing object detection work has considered some object categories as unseen during the training phase and while testing predicts the bounding box for these objects. Like zero-shot object detection, unseen script detection in natural images also needs two essential elements. First is the bounding box regression, and second is identifying the script for the bounding box. At first glance, unseen script detection looks similar to zero-shot object detection. However, from the perspective of bounding box regression, scene text has additional challenges. In scene text, a single character is a valid instance for the bounding box, and a set of characters is also a valid instance for the bounding box. Hence, the primary entity is the character, but the detection requires word-level bounding boxes. It makes bounding box regression for scene text different from the bounding box regression for objects. In some scripts, such as English, the space is considered a hint for word separation. Whereas some other scripts, such as Chinese, is a non-alphabetic language that does not use spaces between words \cite{liu1974effects}. Hence, different scripts have different assumptions for bounding box annotation. Additionally, the similarity of multiple characters with the background and the absence of such scripts during the training phase makes this problem very challenging.

From a script identification perspective, this problem looks similar to an object classification problem. Still, in many aspects, script identification is different from object classification. First, text classification is challenging due to the low-resolution text instances, complex background, and uneven illumination. Second, many scripts share a common set of characters, and only due to the difference of some characters, one script differs from another, such as Greek and English. In contrast, some scripts are entirely different from others based on a holistic view. For example, Chinese and Bangla scripts are perceptually different. To encounter such variability, a model that utilizes the global features and the local stroke information to identify the script is needed. The sequential model has shown their potential for script classification \cite{singh2015can}. In our proposed model, sequential features are used to obtain the sequential dependency of strokes for script identification. At the same time, the global features are used in the proposed system for script identification from a holistic view.

Among different approaches of zero-shot learning, semantic embedding has been widely adopted \cite{norouzi2013zero}. We assume all classes (seen and unseen) must share a common embedding space. The semantic embedding of classes is used as guided information for classification at test time. For semantic embedding, most approaches adopted the word2vec model \cite{mikolov2013distributed}. The same approach has been followed in our work, and word2vec embedding is utilized. The reason to adopt this idea is that in the literature on the writing system, if two language writing styles match, we notice that script names are present together in the text. From this analysis, we have used the word2vec model trained on the writing system corpus as a semantic embedding for our work.

A preliminary version of this work was earlier presented in \cite{keserwani2019zero}. This work has added a significant number of findings and contributions. First, zero-shot script identification is extended to text detection under the zero-shot setting, and a method is presented that shows no need to re-train the model for new scripts. 
Second, the in-depth analysis of the cross-script text detection and understanding of the feasibility of text detection for unseen scripts. The third is the experimental validation of unseen script detection on a public dataset.

\begin{figure*}[!t]
\centering

 \includegraphics[width=\textwidth,height= 0.3\textwidth]{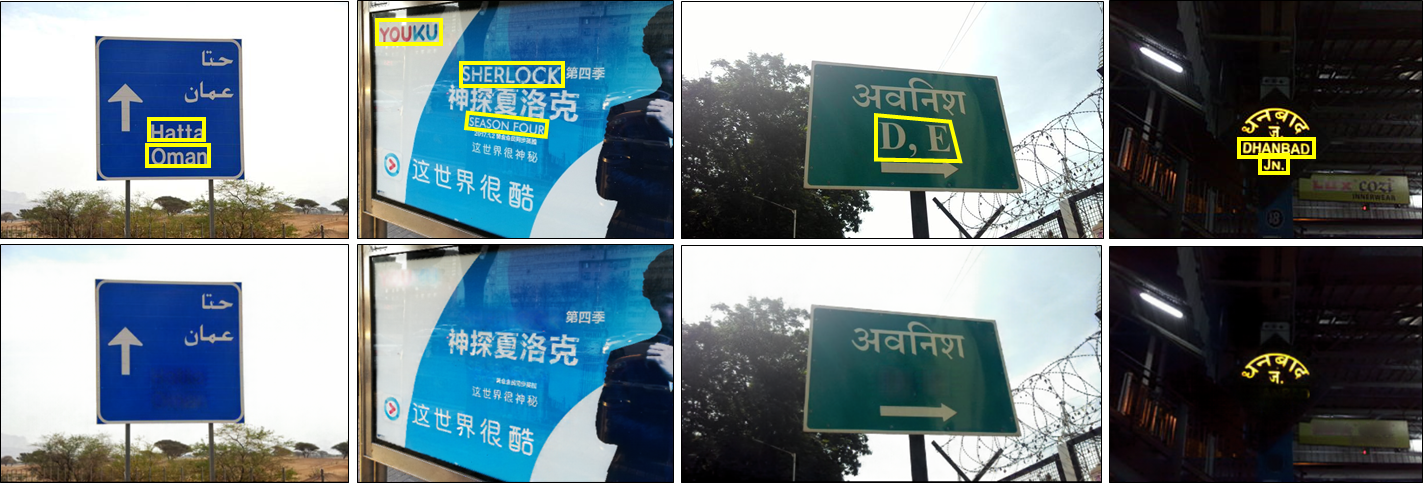}
 
\caption{Script removal from the multi-script images using \cite{keserwani2021text} for generation of the mono-lingual text images. The upper row shows the images with multi-lingual text, whereas the second row shows images with mono-lingual text using the text removal technique. The text region, which is required to be concealed for creating a single script dataset, is shown in yellow rectangles. }
\label{fig:mlt_text_removal}
\end{figure*}

\section{Related Work}
\label{sec:related}
Scene text detection is a well-known computer vision problem and has achieved state-of-the-art performance for most public datasets. Still, for a new script, there is always a need to retrain the model. For retraining, annotation of training data for the new script, is required. In the case of low-resource scripts, data annotation for training is an additional issue. Hence, zero-shot scene text detection is a crucial problem that is still unexplored. However, a few works are present in the literature for zero-shot object detection \cite{bansal2018zero}. 

In contrast to traditional object detection, where only the seen object category is considered a target to be detected, the zero-shot approach detects the unseen object category during testing. In \cite{bansal2018zero}, the authors used visual-semantic embedding for zero-shot detection. They proposed two approaches based on the bounding box generated by edge-boxes \cite{zitnick2014edge}. The first approach was to consider a fixed background class, and the second was the background-aware approach. In \cite{rahman2018zero}, authors proposed to jointly learn the recognition and localization of the object by extending the Faster-RCNN \cite{ren2015faster} in a zero-shot object detection setting. They introduced a loss function that combines max-margin learning and semantic space clustering loss. In \cite{demirel2018zero}, authors proposed a hybrid method by extending YOLOv2 \cite{redmon2017yolo9000}, using a convex combination of embedding and detection. In \cite{rahman2018polarity}, along with joint learning using RetinaNet \cite{lin2017focal}, the authors proposed a polarity loss to align the visual and semantic information through max-margin constraint and refined noisy semantic representation. In \cite{li2019zero}, the authors used the textual description to obtain the semantic embeddings of seen and unseen classes for zero-shot object detection. The authors proposed joint learning of visual-unit and word-level attention. In \cite{rahman2019transductive}, authors used unlabelled unseen class data during training and introduced pseudo-labeling for unlabeled samples. In \cite{zhu2020don}, authors synthesized the visual features based on conditional variational auto-encoder to incorporate unseen object detection. In \cite{zhao2020gtnet}, authors proposed a generative transfer network consisting of an object detector and knowledge transfer module. The knowledge transfer module synthesized the features to generate unseen object features. The IoU-Aware Generative Adversarial Network was used as a feature synthesizer.

\begin{figure*}[!t]
  \centering
  \subfigANDtitle{\includegraphics[width=.19\textwidth]{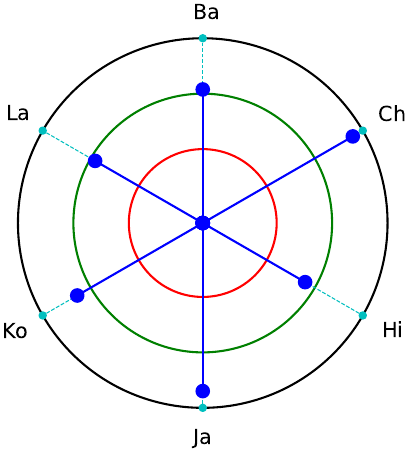} \\ Arabic} \hfill
  \subfigANDtitle{\includegraphics[width=.19\textwidth]{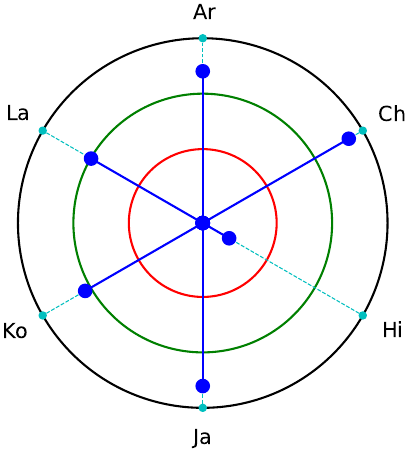} \\ Bangla} \hfill
  \subfigANDtitle{\includegraphics[width=.19\textwidth]{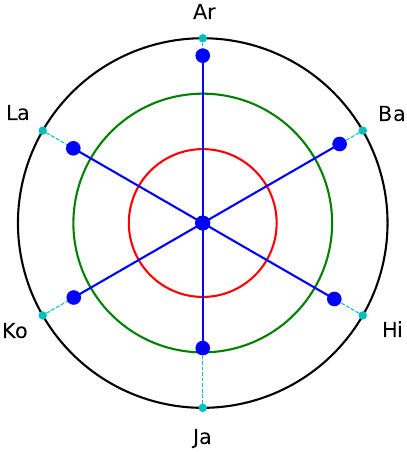} \\ Chinese} \hfill
  \subfigANDtitle{\includegraphics[width=.19\textwidth]{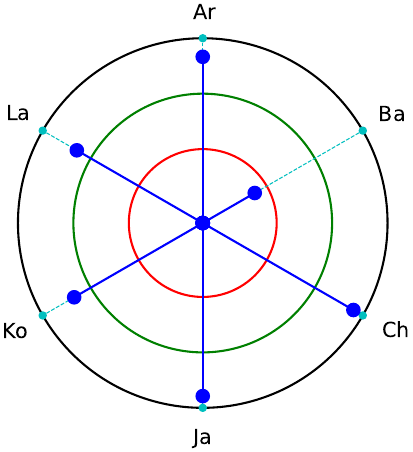} \\ Hindi}

  \medskip

  \hfill
  \subfigANDtitle{\includegraphics[width=.19\textwidth]{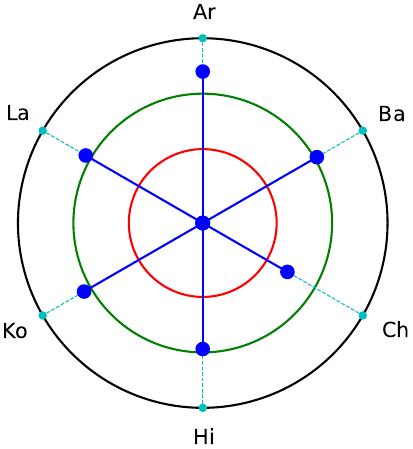} \\ Japanese} \hfill
  \subfigANDtitle{\includegraphics[width=.19\textwidth]{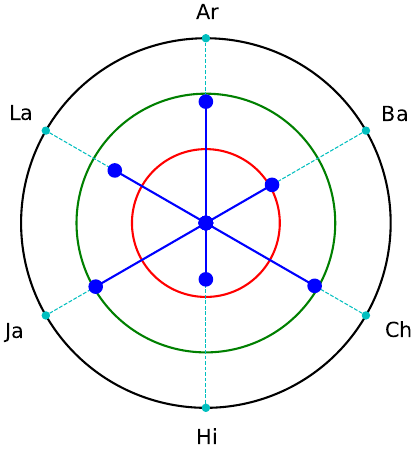} \\ Korean} \hfill
\subfigANDtitle{\includegraphics[width=.19\textwidth]{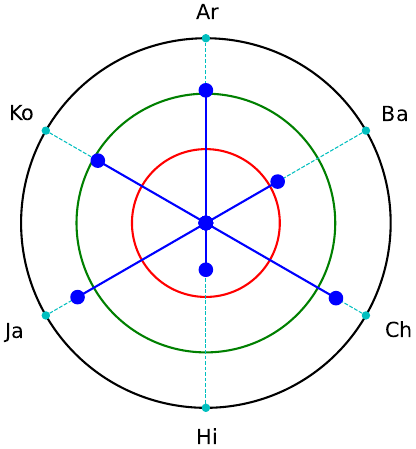} \\ Latin}
  \hfill\mbox{}

\caption{Proximity graph for the closeness of one script with other scripts. The circles with red and green colors represent the closeness of scripts respective to a single script at the f-measure threshold of $0.6$ and $0.3$, respectively. The scripts are aligned with six axes (shown by cyan lines). The shorter lines represent closeness to the respective script. [Annotations: Ar: Arabic, La: Latin, Hi: Hindi, Ba: Bangla, Ko: Korean, Ja: Japanese, Ch: Chinese] }
    \label{fig:closenss}
\end{figure*}

\section{Cross-Script Text Detection} 
\label{sec:cross_script}
Text detection methods trained for a single script or multi-scripts exist in the literature. However, a few fundamental questions require consideration: \emph{Does a text detector achieve enough performance for the cross-scripts? Is it mandatory to train a separate text detector for cross-scripts?} This section tries to find out the answers to the above-raised questions.

\begin{figure*}[!t]
\centering
\subfloat[One Train - Test Rest]{
  \includegraphics[scale=0.5]{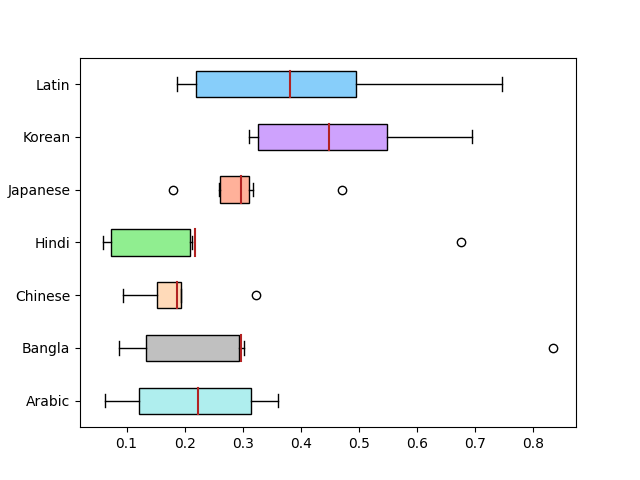}
}
\subfloat[One Test - Train Rest]{
  \includegraphics[scale=0.5]{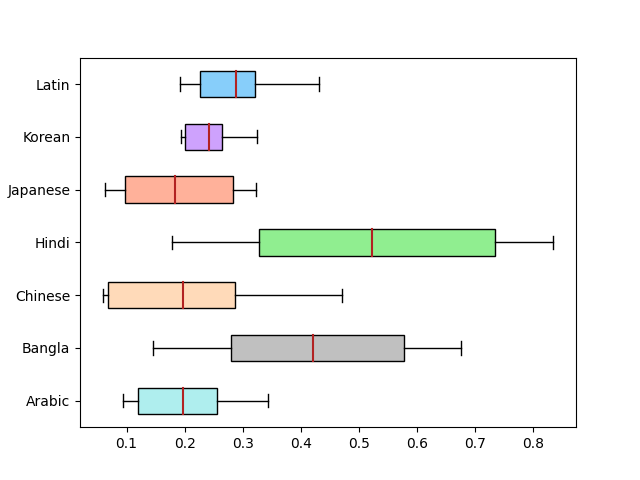}
}

\caption{Box plot to compile the performance of cross-script text detection. The red line marker shows the mean value. (a) Trained on one script (shown on the vertical axis) and tested on the rest of the scripts, (b) Tested on one script (shown on the vertical axis) and trained on the rest of the scripts.  }
\label{fig:cross_script_box_plot_analysis}
\end{figure*}

In text detection, two scripts are considered similar in two ways. Firstly, the visual similarity, in which the character set of one script looks similar to another script. For example, Hindi and Bangla scripts are visually similar. The second criterion for similarity is the nature of bounding boxes for text annotation. Some scripts need word-level bounding boxes, whereas some need line-level bounding boxes. For example, English, Hindi, and Korean scripts need word-level bounding boxes for text annotation, whereas Chinese script needs line-level bounding boxes. These similarities motivate us to investigate the performance of cross-script text detection.

A text detector \cite{keserwani2021quadbox} is chosen for this analysis and is trained on one script and tested on other scripts. A well-known multi-script text detection dataset known as MLT2019 \cite{nayef2019icdar2019} is used to conduct experiments. Seven scripts of the MLT2019 dataset, namely, Arabic, Bangla, Chinese, Hindi, Japanese, Korean, and Latin, are considered. Since English is a script used internationally, it is present with other scripts in many images. Hence, we have used the text removal technique \cite{keserwani2021text} to convert the multi-script text images to single-script. Some sample images before the text removal and after text removal for the cross-script analysis are shown in Fig. \ref{fig:mlt_text_removal}. For the analysis purpose, we considered the f-measure \cite{ye2014text} metric. The closeness of scripts is shown with the proximity graphs (Fig. \ref{fig:closenss}). The proximity graph presents a visual way to show the closeness of a script with other scripts. The cross-scripts are represented by axes (cyan-colored lines). The closest cross-script has the smallest line. 

Two f-measure thresholds are utilized for this analysis. The first threshold is set to $0.6$, which shows that the scripts are similar, and cross-script text detection can be recommended for such a set of scripts. The analysis shows that the model trained on Latin script can be used directly for the Bangla and Hindi scripts during inference. Likewise, the model for Hindi script can be used for evaluating Bangla script and vice-versa. The second f-measure threshold is set to $0.3$. It helps to find scripts that are similar but loosely coupled. Chinese and Japanese scripts fall into this category. The behavior of Korean script demonstrates its closeness with Hindi, Bangla, and Latin scripts more than visually similar Japanese and Chinese scripts. The reason is a difference in the nature of the bounding box annotation of Korean script from Chinese and Japanese scripts. The Chinese and Japanese scripts use line-level bounding box annotation, and the Korean script uses word-level bounding box annotation in the ground truth. Hence, the model trained on Korean script favors the word-level bounding boxes while evaluating Hindi, Bangla, and Latin scripts. Therefore, despite the visual similarity of Korean with Japanese and Chinese scripts, the model trained in Korean script shows better text detection for Bangla, Hindi, and Latin scripts. The proximity graph also reveals that the Arabic script is close to Hindi and Latin scripts. 

To further analyze the distribution of cross-script text detection performance, the box plots of f-measures of the text detectors are shown in Fig. \ref{fig:cross_script_box_plot_analysis}. In Fig. \ref{fig:cross_script_box_plot_analysis} (a), the box plot shows the distribution of the performance of a model trained on a single script and tested on the rest of the scripts. From Fig. \ref{fig:cross_script_box_plot_analysis}(a), it is observed that the model trained in Korean script has the best mean f-measure among others. In contrast, the Chinese script has the worst mean f-measure. The Hindi script has Bangla as an outlier and vice-versa, as shown by the dotted points. In the case of Bangla script, other than Hindi, the performance of the rest of the scripts is far inferior. The same trend is observed in Hindi. Chinese and Japanese scripts have shown similar behavior to Hindi and Bangla scripts. In Fig \ref{fig:cross_script_box_plot_analysis} (b), the box plot shows the distribution of performance on a single script by models trained on the rest of the scripts. Among them, the mean f-measure of Hindi is the best, and Arabic is the worst.

This cross-script text detection analysis helps to lay the foundation for unseen script detection. The cross-script text detection demonstrates the capability of a text detector without getting trained on or fine-tuned to a target script. From this analysis, it has been concluded that cross-script text detection is feasible for similar scripts. Based on the above analysis, the following section proposes a baseline method for unseen script detection in scene images.

\section{Unseen Script Detection} 
\label{sec:zstd}

Suppose the text detection for multi-scripts is composed of $n$ scripts and $n=n_0+n_1$, where $n_0$ is the number of classes for seen scripts during training, and $n_1$ is the number of the classes for the unseen scripts during training. Under the hypothesis of the paper, only the bounding box of the seen scripts is provided during the training phase, whereas testing is performed on the unseen scripts. For an image $I \in \mathbb{R}^{P \times Q \times 3}$, there are $b$ number of bounding boxes $B=\{b_i| i \in \{1,...|b|\}\}$ with scripts $C=\{y_i| i \in \{1,...|b|\}\}$.   

The fundamental approach adopted for unseen script detection in natural images consists of two parts, namely, (a) cross-script bounding box prediction and (b) unseen script identification. The block diagram for the proposed baseline approach for unseen script detection is shown in Fig. \ref{fig:zero_shot_text_detection}. The proposed approach has three basic building blocks. These are:

\begin{enumerate}

    \item \emph{Bounding Box Prediction :} Based on the cross-script analysis, the prediction of the bounding box for unseen categories can be made using an on-the-shelf text detector. In this work, Quadbox text detector \cite{keserwani2021quadbox} has been used. The Quadbox text detector has been trained on the seen category of scripts. The Quadbox returns the set of bounding boxes $\hat{B}=\{b_i|i\in(1,\hat{n})\}$. The text detector generates the bounding box irrespective of the script categories.  
    
    \begin{figure}[!t]
    \centering
    \includegraphics[ width=0.48\textwidth]{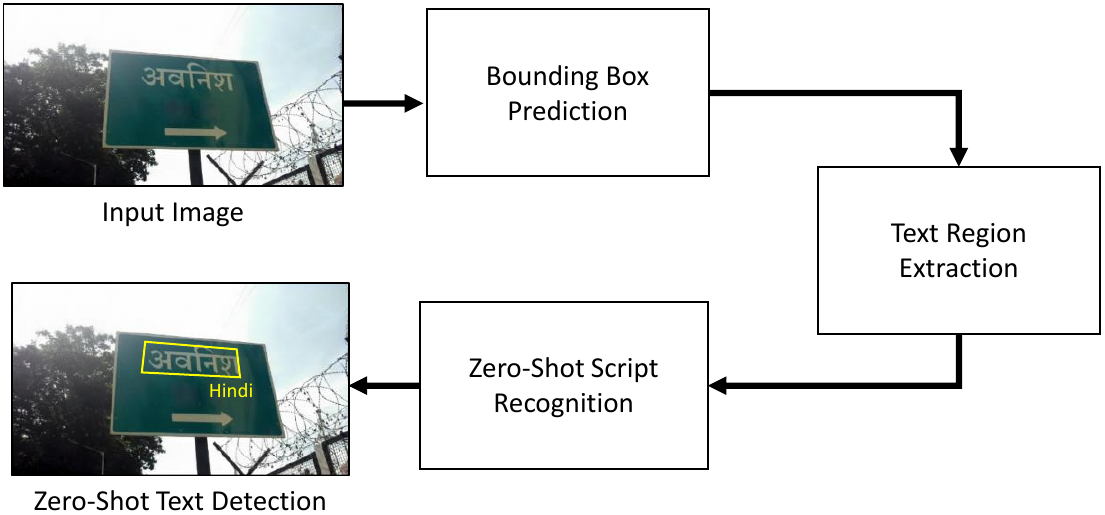}
    \caption{Fundamental pipeline for the unseen script detection. }
    \label{fig:zero_shot_text_detection}
\end{figure}

    \item \emph{Text Region Extraction :} The bounding boxes generated from the bounding box prediction phase are cropped out from the image in the text region extraction phase. The bounding box predicted by the bounding box prediction phase consists of four coordinates $b_t=\{(x_i,y_i)|i\in (1,4)\}$. A minimum area rectangle has been computed for the bounding box represented by the four coordinates. Then the image is rotated with the angle of the minimum area rectangle. The rotated image rectangle is cropped for the text region extraction.
    
    \item \emph{Unseen Script Recognition :} The extracted text regions $R$ are considered as the candidate regions for script recognition. The unseen script identification uses the model trained on seen script categories and identifies the category of text region $r_t$. The unseen script identification part is taken from \cite{keserwani2019zero}.
    
\end{enumerate}

\section{Experiment}
\label{sec:experiment}
\subsection{Dataset Description}
In scene text detection, a well-known multi-script dataset known as the MLT2019 dataset \cite{nayef2019icdar2019} consists of seven scripts, namely Chinese, Japanese, Korean, Latin, Arabic, Bangla, and Hindi. In addition, this dataset has some 'symbols' and 'mixed' categories. Seven scripts mentioned above are considered for this work. For unseen script detection, a subset of the seen and unseen script categories from the zero-shot script identification has been taken as the seen and unseen script categories from the MLT2019 dataset. Hence, from the MLT2019 dataset, the seen and unseen categories for unseen script detection are \{Arabic, Latin, Bangla, Japanese\} and \{Chinese, Korean, Hindi\}, respectively. However, some images have scripts from both seen and unseen categories. Hence, we consider only those images that contain scripts from seen or unseen categories. The statistics of the dataset used for this work are given in Table \ref{table:dataset_zstd}.

\begin{table}[!t]
\centering
\caption{Statistics of the used dataset for the unseen script detection. }
\begin{tabular}{|c|l|c|}
\hline
\textbf{Category} & \textbf{Scripts}  &  \textbf{\#Images} \\ \hline
\multirow{4}{*}{Seen} & Latin & 3996 \\ \cline{2-3}
                      & Bangla & 1000 \\ \cline{2-3}
                      & Arabic & 1000 \\ \cline{2-3}
                      & Japanese & 940 \\ \hline
\multirow{3}{*}{Unseen} & Chinese & 454\\ \cline{2-3}
                        & Korean   & 241 \\ \cline{2-3}
                        & Hindi    & 816  \\ \hline
\end{tabular}
\label{table:dataset_zstd}
\end{table}

\subsection{Implementation Details}
The proposed model for unseen script detection requires two separately trained models. It requires one model for script identification and one for cross-script text bounding box regression. Both models are implemented in Pytorch\footnote{https://pytorch.org/} and trained on GTX 1070 Ti GPU. 
For the unseen script detection in natural images, the text detector  \cite{keserwani2021quadbox} is trained on the unseen category for $400$ epochs with a batch size of $4$. The model was trained with an initial learning rate of $10^{-4}$, and the learning rate decays by $0.1$ after $200$ epochs. For training, the data augmentations used are image rotation, image scaling, and random cropping. During inference, vanilla non-maximal suppression post-processing has been employed to filter out the bounding boxes. For unseen script identification, we have taken the trained model of the paper \cite{keserwani2019zero} on which this paper is extended.

\subsection{Evaluation Protocol}

\begin{figure}[!t]
\includegraphics[width=0.45\textwidth]{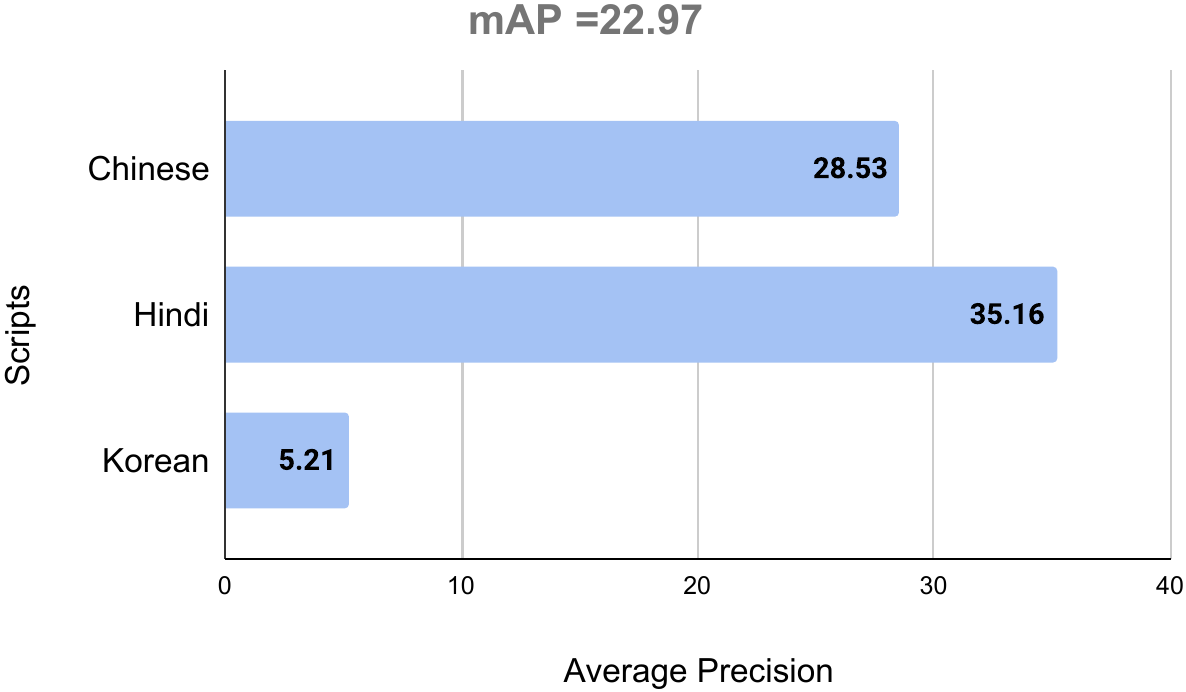}
\centering
\caption{Quantitative performance of the unseen script detection.}
\label{Fig:zstd_performance}
\end{figure}

\begin{figure}[!t]
\includegraphics[width=0.45\textwidth]{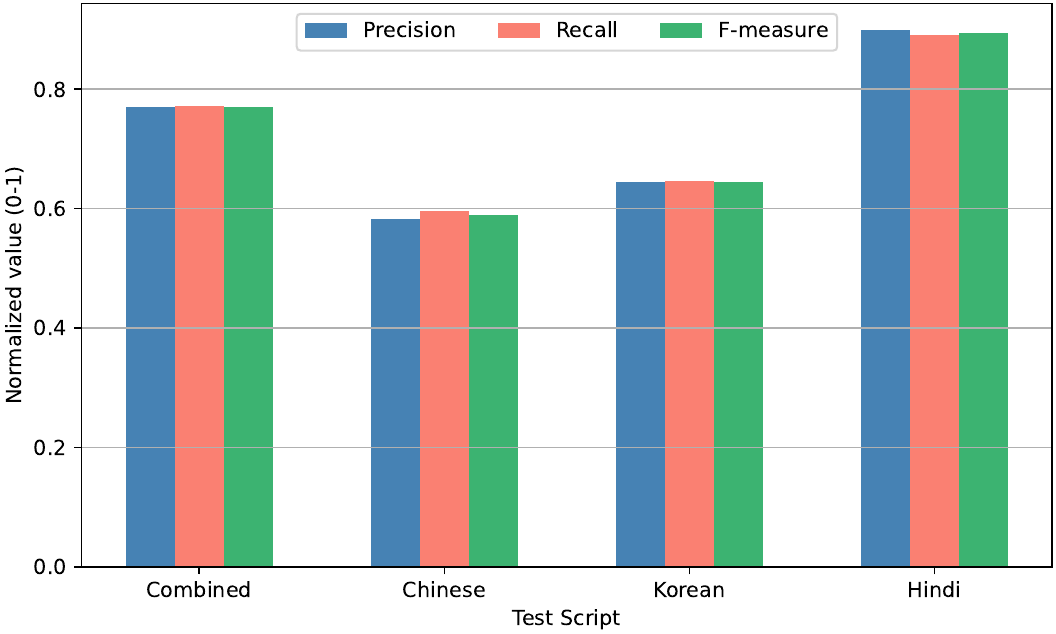}
\centering
\caption{Performance of bounding box regression by unseen script detection. The model performance has been computed over a combined image set of Chinese, Korean, and Hindi scripts (Combined) and individual scripts (Chinese, Korean, and Hindi).}
\label{Fig:zstd_fm_performance}
\end{figure}

The same protocol as in zero-shot object detection~\cite{everingham2015pascal} has been followed for the unseen script detection to evaluate natural images. The recall is defined as the ratio of all positive examples rated above a given threshold. Precision is the ratio of all examples above that rank from the positive class. The precision/recall curve is computed from a method’s ranked output for a given task and class. The average precision (AP) summarises the shape of the precision/recall curve and is defined as the mean precision at eleven equally spaced recall levels [0, 0.1,..., 1]. The average precision (AP) has been computed for each unseen category, and subsequently, the overall performance has been computed using the mean average precision (mAP) metric.

\begin{figure*}[!t]
\includegraphics[width=0.99\textwidth]{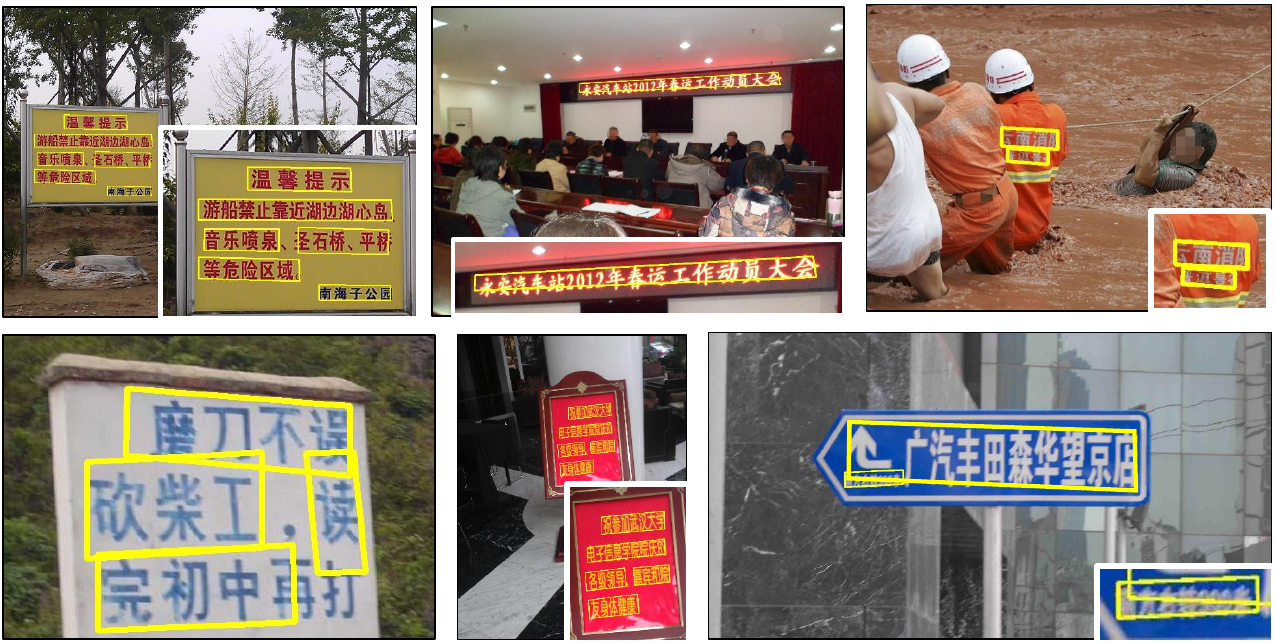}
\centering
\caption{Unseen Chinese script detection in natural images. The yellow color bounding box shows the Chinese script. The smaller text regions are cropped out and shown at the corner of each image at a higher resolution for better visualization.  }
\label{Fig:chinese_results}
\end{figure*}

\begin{figure*}[!t]
\includegraphics[width=0.99\textwidth,height=8.0cm]{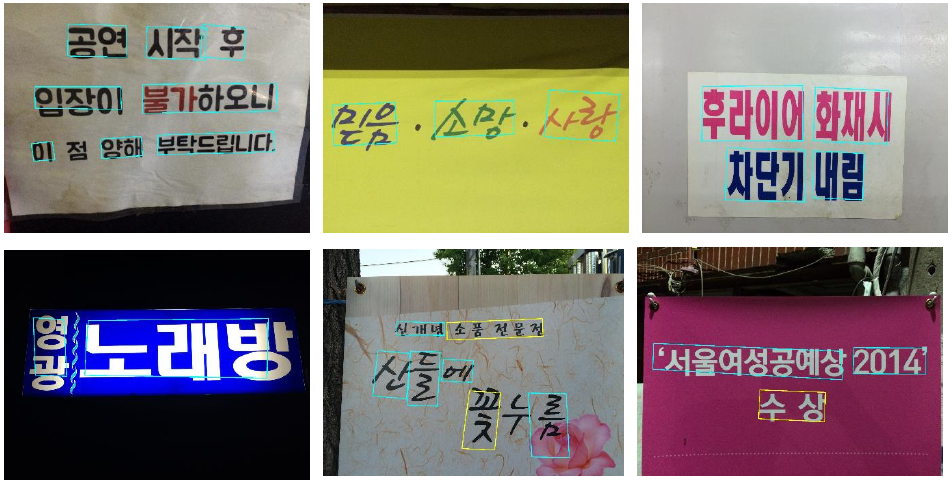}
\centering
\caption{Unseen Korean script detection in natural images. The cyan color bounding box shows the Korean script. The yellow color bounding box shows the Chinese script, which is miss-classified. }
\label{Fig:korean_results}
\end{figure*}

\begin{figure*}[!t]
\includegraphics[width=0.99\textwidth,height=7.0cm]{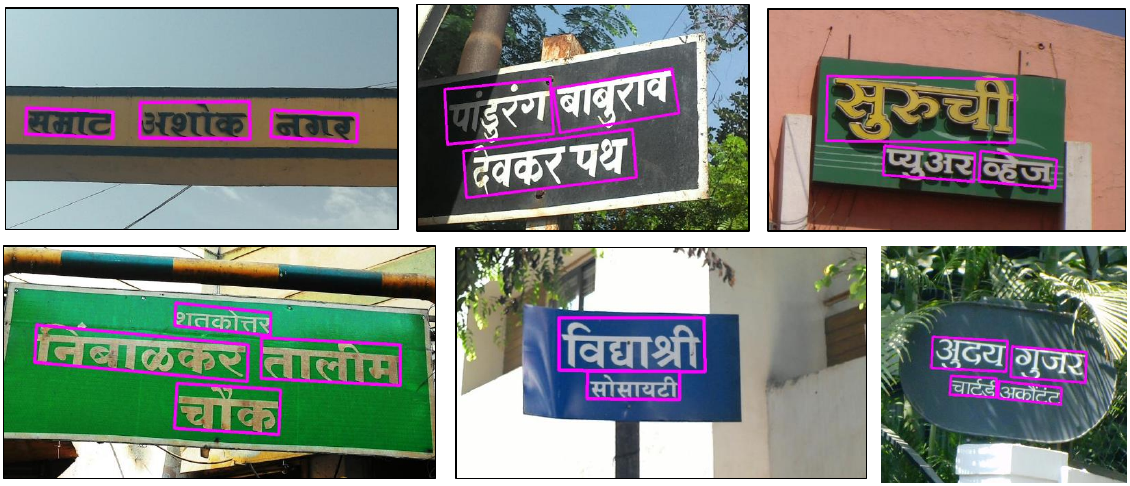}
\centering
\caption{Unseen Hindi script detection in natural images. The pink color bounding box shows the Hindi script.}
\label{Fig:hindi_results}
\end{figure*}

\subsection{Quantitative Results} 
Two factors decide the performance of unseen script detection: (a) the classification performance to identify the cross-scripts, (b) predicting bounding box which possesses sufficient overlap with the ground truth (in this experiment, the intersection over the union of $0.5$ has been taken into consideration). The quantitative performance of the unseen script detection in natural images is shown in Fig. \ref{Fig:zstd_performance}. These results show the combined impact of the classification and box regression. The achieved performance for the proposed pipeline is $22.97\%$ mAP. Hindi script is the best-performing among Chinese, Korean, and Hindi scripts. From the perspective of unseen script (zero-shot) recognition, the visual similarity among the scripts plays an important role. Bangla and Hindi scripts have a close visual appearance. Bangla script has been used in the training phase, and it helps to achieve good performance for Hindi in the testing phase. Similarly, the visual similarity of Japanese with Chinese and Korean scripts should have helped their recognition. However, the visual closeness of Chinese and Korean scripts hinders the classification performance.

Another perspective is the proper bounding box regression. Bounding box regression performance has been computed using well-known precision, recall, and f-measure text detection metrics to understand this perspective. This bounding box performance evaluation has been conducted in two different ways. Firstly, the bounding box performance evaluation has been done for combined scripts (i.e., Chinese, Hindi, and Korean). Secondly, it has been done for individual scripts. The precision, recall, and f-measure for both ways are summarized in Fig. \ref{Fig:zstd_fm_performance}. From the Fig. \ref{Fig:zstd_fm_performance}, it has been observed that combined scripts testing has achieved the precision, recall, and f-measure of $0.7701$, $0.7709$, and $0.7705$, respectively. These results show that the cross-script bounding box prediction is plausible. Bounding box performance evaluation for individual scripts shows that the model has performed best for Hindi. The performance of Hindi script is better due to the high visual similarity between Hindi and Bangla scripts. Since Bangla is used in training, it helps the model predict good bounding boxes for Hindi during testing. However, for the Chinese, the model performance is comparatively inferior. The need for line-level bounding box prediction performance is the reason behind the low performance of the model trained on the Chinese script. Since most training scripts have word-level bounding box ground truth, the model is more inclined toward predicting word-level bounding box regression.  

\subsection{Qualitative Results} 
The bounding box predicted along with the script categories are visually shown in Fig. \ref{Fig:chinese_results}, Fig. \ref{Fig:korean_results}, and Fig. \ref{Fig:hindi_results}. The Chinese, Korean, and Hindi bounding boxes are shown in qualitative visualization with yellow, cyan, and pink colors. From Fig. \ref{Fig:hindi_results} and Fig. \ref{Fig:korean_results}, it has been observed that the bounding box appears much tighter and more appropriate. It is due to the word-level bounding box requirement of Korean and Hindi. In Fig. \ref{Fig:chinese_results}, it has been observed that the predicted bounding box appears shorter in length due to the requirement of the line-level bounding box for the Chinese script. It results in a lower value of the f-measure (shown in Fig. \ref{Fig:zstd_fm_performance}) than the scripts that need word-level bounding boxes, such as Hindi and Korean. This observation also correlates with the f-measure values for Korean and Hindi (shown in Fig. \ref{Fig:zstd_fm_performance}).

\subsection{Comparative Analysis} 
A comparision has been performed by combining the bounding box prediction model with various zero-shot script identification models. The proposed pipeline has been computed by combining the quadbox (QB) with four variants of zero-shot script identification models (mentioned in \cite{keserwani2019zero}). These zero-shot script identification models are (a) VGG-16 + Global Average Pooling (GAP) [BL1], (b) VGG-16 + LSTM [BL2], and (c) VGG-16 + SEB + GAP [V]. Among them, the proposed pipeline with SEB and GAP is the best performing. The ablation is summarized in Table \ref{Table:comaprison_zstd}.

\begin{table}[!t]
\centering
\caption{Comparative analysis of the proposed pipeline with various zero-shot script identification methods.}
\begin{tabular}{|l|lll|l|}
\hline
\multirow{2}{*}{\textbf{Methods}} & \multicolumn{3}{c|}{\textbf{Average Precision}}                             & \multirow{2}{*}{\textbf{mAP}} \\ \cline{2-4}
                         & \multicolumn{1}{c|}{\textbf{Chinese}} & \multicolumn{1}{c|}{\textbf{Hindi}} & \textbf{Korean} &                      \\ \hline
QB + BL1  &  \multicolumn{1}{c|}{19.59} & \multicolumn{1}{c|}{28.43} &  \multicolumn{1}{c|}{6.03}      &  18.02                    \\ \hline
QB + BL2  & \multicolumn{1}{c|}{23.85}  & \multicolumn{1}{c|}{28.93}   &  \multicolumn{1}{c|}{4.95}    &    19.24\\ \hline

QB + V  &  \multicolumn{1}{c|}{23.03}    & \multicolumn{1}{c|}{27.01}   &     \multicolumn{1}{c|}{4.56}  & 18.20     \\ \hline
QB + SEB + GAP   &  \multicolumn{1}{c|}{28.53} &  \multicolumn{1}{c|}{35.16} &  \multicolumn{1}{c|}{5.21}      &    \textbf{22.97}   \\ \hline
\end{tabular}
\label{Table:comaprison_zstd}
\end{table}

\section{Conclusion}
\label{sec:conclusion} 
In this work, the analysis to understand the impact of text detectors on cross-script is performed. The analysis concludes that an on-the-shelf text detector can be used for unseen scripts that are either visually similar or possess similar annotation level requirements (i.e., word-level or line-level). This analysis helps to develop the fundamental pipeline for the zero-shot learning approach for unseen script detection. It questioned the need for separate training for unseen scripts. With the combination of the cross-script text bounding box prediction and unseen (zero-shot) script identification, unseen script detection can be accomplished. The proposed method is highly dependent on the effectiveness of the word2vec model, which can be improved in the future by using a better and more extensive corpus.   

\bibliographystyle{IEEEtran}
\bibliography{ref}

\end{document}